\documentclass{article}

\PassOptionsToPackage{numbers, compress}{natbib}
 \usepackage[preprint]{neurips_2026}

\usepackage[utf8]{inputenc} % allow utf-8 input
\usepackage[T1]{fontenc}    % use 8-bit T1 fonts
\usepackage{hyperref}       % hyperlinks
\usepackage{url}            % simple URL typesetting
\usepackage{booktabs}       % professional-quality tables
\usepackage{amsfonts}       % blackboard math symbols
\usepackage{nicefrac}       % compact symbols for 1/2, etc.
\usepackage{microtype}      % microtypography
\usepackage{xcolor}         % colors
\usepackage{enumitem}
\usepackage{graphicx}
\usepackage{amsmath}
\usepackage{colortbl}
\usepackage{multirow}
\usepackage{multicol}
\usepackage{wrapfig}
\usepackage{caption}
\usepackage{pifont}
\usepackage[most]{tcolorbox}
\usepackage{makecell}
\title{video-SALMONN-R$^3$: Learning to ReWatch, ReAsk, and ReAnswer for Efficient Video Understanding}

\author{%
  Yixuan Li$^1$, Guangzhi Sun$^3$, Yudong Yang$^1$, Chao Zhang$^1$ \\
  $^1$Tsinghua University \quad $^2$University of Cambridge\\
  \texttt{liyixuan25@mails.tsinghua.edu.cn}\quad \texttt{cz277@tsinghua.edu.cn} \\
}

\begin{document}

\maketitle

\begin{abstract}
Video large language models (LLMs) are often constrained by computation and memory budgets, leading them to use reduced frame rates and spatial resolutions, which may cause them to miss critical information for question answering (QA). A practical and efficient solution is a two-stage paradigm: first perform coarse video understanding to localize relevant segments, and then re-watch these segments at higher temporal or spatial fidelity.
In this paper, we present video-SALMONN-R$^3$, the first end-to-end video-LLM that enables \textbf{re-watch} through reinforcement learning without relying on chain-of-thought (CoT) cold-start. This design removes the need for costly CoT data annotations and avoids CoT-based supervised fine-tuning (SFT), which can otherwise degrade the pretrained video understanding abilities.
To address the mismatch between the reasoning-first behavior induced by re-watch and the answer-first tendency of pretrained video-LLMs, we propose a \textbf{re-answer} strategy, in which the model first produces a direct answer in the first watch and then refines it after re-watching. Finally, to improve question adherence during re-watching, we propose a \textbf{re-ask} mechanism that re-injects the query when revisiting localized segments.
Experimental results show that video-SALMONN-R$^3$ consistently outperforms both the base model and the QA-SFT baseline, while surpassing prior re-watch-based approaches with significantly lower computational cost. Code, models, and data will be publicly released upon acceptance.
  %We present video-SALMONN-R$^3$, the first end-to-end video large language model that acquires a \textbf{re-watch} temporal-localization capability purely through reinforcement learning, without relying on any chain-of-thought cold-start annotations. Instead of curating costly interleaved reasoning-and-action CoT corpora to bootstrap localization behavior, which is likely to degrade performance on general video understanding, we start from an instruction-tuned base model and directly apply reinforcement learning to acquire the re-watch ability. To preserve the same level of performance as the original model, we introduce a \textbf{re-answer} mechanism, where the model first answers from its well-aligned prior and then refines the answer after temporal grounding. We further propose a \textbf{re-ask} mechanism: by re-posing the question after the localized segment is revisited, the model strengthens question-video interaction in the second pass under causal attention. As a result, video-SALMONN-R$^3$ delivers consistent improvements over the base model and QA-SFT baseline, and outperforms prior localization-based methods while using noticeably fewer computational resources. Our source code, models, and data will be released.
\end{abstract}

\vspace*{-0.1cm}
\section{Introduction}\label{sec:intro}
\vspace*{-0.1cm}
With rapid advances in multimodal large language models (LLMs), their capabilities have expanded from text-only understanding to images, audio, and, more recently, video \cite{li2024llavaov, zhang2024video, tang2024salmonn, tang2025video, zhang2025videollama, chen2025avocado, xu2025qwen3, bai2025qwen3}.
Among these modalities, video presents unique challenges due to its high dimensionality and long temporal horizon. A single high-resolution, long-duration video can generate an extremely large number of visual and acoustic tokens, making exhaustive processing infeasible under current computational and memory constraints. To ensure tractability, most existing video-LLMs rely on sparse frame sampling or aggressive token compression \cite{liu2024nvila, li2025improving, qin2025video, sun2025video}. While effective at reducing computational cost and capturing coarse global context, these strategies can limit access to fine-grained temporal and semantic details, which are critical for tasks requiring precise spatiotemporal reasoning beyond holistic understanding.

%Large language models (LLMs) have progressed remarkably in recent years, evolving from simple chatbots into practical tools used across a wide range of industries \cite{openai2024gpt4technicalreport, comanici2025gemini, guo2025deepseek, team2026kimi, zeng2026glm, qwen3}. Building on this progress, multimodal LLMs have rapidly expanded the frontier, extending language understanding to images, audio, and eventually video \cite{li2024llavaov,zhang2024video,tang2024salmonn,tang2025video,zhang2025videollama,chen2025avocado,xu2025qwen3,bai2025qwen3}. Among these modalities, video understanding stands out as one of the most difficult frontiers. A single video is inherently a long, dense sequence of visual and acoustic signals, and naively feeding all of its frames into an LLM produces an overwhelming number of multimodal tokens, far beyond what current architectures can afford to process. To keep computation tractable, most existing methods resort to sparse frame sampling or aggressive token compression \cite{liu2024nvila,li2025improving,qin2025video,sun2025video}. While practical, both strategies inevitably discard fine-grained temporal and semantic cues, and the resulting information loss quickly becomes a bottleneck once questions demand precise reasoning over specific moments rather than a global gist of the video.

To address this limitation, recent work has explored a two-stage \emph{re-watch} paradigm: the model first performs coarse video understanding to identify relevant segments, and then revisits these segments at higher temporal or spatial fidelity. This design separates localization (“where to look”) from reasoning (“what to answer”), and has shown clear advantages over single-pass pipelines. However, existing approaches suffer from two key drawbacks. One line of work decomposes the process into multiple specialized models that communicate via natural language \cite{zuo2025videolucy,yeo2025gcagent}. Although flexible, such systems incur substantial computational overhead and limit reasoning fidelity, as the final decision-making module lacks direct access to raw visual evidence. Another line embeds re-watch behavior within a single model via multi-pass reasoning trajectories \cite{fu2025love,yang2025longvt,yanvideochat,li2025videochat,ding2025videozoomer,chen2026think}. While architecturally cleaner, this approach depends heavily on large-scale interleaved reasoning-and-action chain-of-thought (CoT) annotations, which are expensive to construct and often introduce undesirable biases that degrade the pretrained model’s original video understanding capabilities \cite{feng2025video}.

Motivated by these limitations, we ask whether temporal localization can be incorporated into a strong video-LLM without relying on costly CoT cold-start annotations and without sacrificing existing performance. We answer this affirmatively with video-SALMONN-R$^3$, an end-to-end video LLM that acquires a \textbf{re-watch} capability purely through reinforcement learning (RL). Starting from an instruction-tuned base model, we directly optimize re-watch behavior using RL, avoiding both CoT annotation and CoT-based supervised fine-tuning (SFT).
A central challenge in this setting is the mismatch between the reasoning-first behavior induced by re-watch and the answer-first tendency of pretrained video-LLMs. To address this, we introduce a \textbf{re-answer} strategy, in which the model first produces an answer based on its well-aligned prior during the first pass, and then refines this answer after re-watching the localized segments. %This design anchors the newly acquired localization ability to existing capabilities, enabling the model to improve without forgetting its prior knowledge.
This design anchors the newly acquired localization ability to existing capabilities, enabling the model to improve without overwriting its prior knowledge.
%
%Furthermore, we identify a limitation arising from causal attention: the question tokens in the first pass cannot attend to frames observed during re-watching, weakening question–video interaction in the second stage. To resolve this, we propose a \textbf{re-ask} mechanism that re-injects the query when revisiting localized segments, allowing direct interaction between the question and the newly observed evidence. This simple yet effective modification leads to consistent performance gains without additional computational cost.
Furthermore, we identify a limitation arising from causal attention: the question tokens in the first pass cannot attend to frames observed during re-watching, weakening the question adherence in the second stage. To resolve this, we propose a \textbf{re-ask} mechanism that re-injects the query when revisiting localized segments, allowing direct interaction between the question and the newly observed audio-visual evidence. This simple yet effective modification leads to consistent performance gains with little additional computational cost.

Our contributions are summarized as follows:

\begin{itemize}[itemsep=0pt, leftmargin=*]
%\item We present video-SALMONN-R$^3$, the first end-to-end video LLM that acquires re-watch temporal localization purely via dynamic sampling policy optimization (DAPO)\cite{yudapo} based RL on an instruction-tuned base model, eliminating the need for CoT cold-start annotations while achieving superior performance with lower computational cost.
\item We present video-SALMONN-R$^3$, the first end-to-end video LLM that acquires re-watch temporal localization purely via dynamic sampling policy optimization (DAPO) \cite{yudapo} based RL on an instruction-tuned base model, eliminating the need for CoT cold-start annotations while achieving new state-of-the-art (SOTA) performance on six benchmarks with lower computational cost.
\item We propose a re-answer strategy that reconciles the mismatch between re-watch-induced reasoning and the pretrained model’s answer-first behavior, enabling stable integration of localization capability without degrading existing performance.
\item We introduce a re-ask mechanism that re-injects the query during re-watching, addressing a causal-attention limitation and strengthening question–video interaction with negligible overhead.
\end{itemize}

\vspace*{-0.1cm}
\section{Related Works}

\vspace*{-0.1cm}
\subsection{Multimodal LLMs for Video Understanding}
\vspace*{-0.1cm}

Recent video large language models (video-LLMs) typically connect a visual encoder to an LLM via a lightweight projector, representing videos as sparsely sampled frames, as seen in LLaVA-Video \cite{zhang2024video}, VideoLLaMA 3 \cite{damonlpsg2025videollama3}, Qwen3-VL \cite{bai2025qwen3}, Video-Chat-Flash \cite{li2024videochat}, and Intern-VL 3.5 \cite{wang2025internvl3_5}. A growing body of work further incorporates audio to enable joint audio-visual reasoning, including VideoLLaMA 2 \cite{damonlpsg2024videollama2}, video-SALMONN \cite{sun2024videosalmonn,tang2025video}, Qwen2.5-Omni and Qwen3-Omni \cite{xu2025qwen2,xu2025qwen3}, AVoCaDo \cite{chen2025avocado}, Omni-Captioner \cite{omni-captioner}, D-ORCA \cite{tang2026dorca}, and MiniCPM-o 4.5 \cite{cui2026minicpm}.

Despite these advances, such models remain constrained by memory and computational budgets, and thus typically operate at reduced frame rates and spatial resolutions, limiting access to fine-grained visual details. To alleviate this, prior work has explored a range of complementary strategies. Methods such as NVILA \cite{liu2024nvila}, Tempo \cite{fei2026small}, and F-16 \cite{li2025improving} employ multi-frame fusion to reduce token redundancy, while Video-XL-2 \cite{qin2025video}, AdaReTaKe \cite{wang2025ada}, and APVR \cite{gao2026apvr} adopt chunk-wise prefilling for long-video compression. Other approaches, including VidCom$^2$ \cite{liu2025vidcom2}, DyTo \cite{zhang2024beyond}, and InfiniPot-V \cite{kim2025infinipot_v}, explore training-free token compression schemes, whereas VideoStreaming \cite{qianstreaming}, video-SALMONN S \cite{sun2025video}, Flash-VStream \cite{zhang2024flashvstream}, and Dispider \cite{qian2025dispider} maintain streaming memories to capture long-term dependencies.
While effective, these approaches typically place substantial responsibility on the projector for information compression or rely on heuristic criteria such as attention scores or feature similarity. More recently, a re-watching paradigm has emerged, in which the model processes the video iteratively and adaptively determines frame sampling strategies across passes. By leveraging the LLM’s intrinsic reasoning abilities, this paradigm reduces reliance on manually designed heuristics and offers a promising direction for more effective video understanding.

\vspace*{-0.1cm}
\subsection{Video Re-Watching by Temporal Localization}
\vspace*{-0.1cm}

Existing approaches to temporal localization in video-LLMs generally follow two design paradigms, differing in whether the re-watch process is implemented at the system level via multiple agents or within a single unified model.

\textbf{Multi-agent approach.} One line of work realizes re-watch through multi-agent collaboration. These frameworks typically compress long videos into textual memories and enable a reasoning agent to iteratively localize and revisit query-relevant segments, often processing them at higher temporal or spatial fidelity. For example, VideoLucy \cite{zuo2025videolucy} organizes memory into a hierarchical structure with progressively finer granularity and performs agent-based backtracking from coarse to fine, while GCAgent \cite{yeo2025gcagent} constructs schematic and narrative episodic memories from speech transcripts and couples a memory manager with a reasoning agent within a perception–action–reflection loop. While the modular design offers flexibility, coordinating multiple LLMs can introduce substantial computational overhead. Moreover, as the final answering agent typically operates on textual representations rather than raw video inputs, its reasoning often depends on intermediate summaries, which may limit access to certain fine-grained audio-visual details.

\textbf{Single-model approach.} Another line of work integrates the entire re-watch loop within a single video-LLM, alternating between global skimming and fine-grained inspection across multiple passs. A common approach constructs CoT trajectories that interleave textual reasoning with temporal actions, initializes the base model via SFT, and subsequently refines the agentic policy through RL. For instance, VideoChat-R1.5 \cite{yanvideochat, li2025videochat} formulates this as visual test-time scaling with iterative perception over high-confidence spatiotemporal regions; LOVE-R1 \cite{fu2025love} combines dense low-resolution global frames with adaptive high-resolution refinement under a CoT SFT followed by a decoupled RL stage; LongVT \cite{yang2025longvt} models re-watching as tool-based clip retrieval on demand; and VideoZoomer \cite{ding2025videozoomer} employs a temporal localization and revisiting operator within a multi-pass SFT-then-RL pipeline. While these approaches provide an elegant end-to-end formulation, they typically rely on a CoT-based cold-start stage. Constructing high-quality interleaved CoT trajectories can be labor-intensive and difficult to scale, and the resulting supervision may introduce distributional or stylistic biases that influence the pretrained model’s original behavior \cite{feng2025video}.

\vspace*{-0.1cm}
\section{Methods}
\vspace*{-0.1cm}
%\subsection{Model Pipeline}
\subsection{Model Workflow}
\vspace*{-0.1cm}

\begin{figure}[t]
    \centering
    \includegraphics[width=0.9\linewidth]{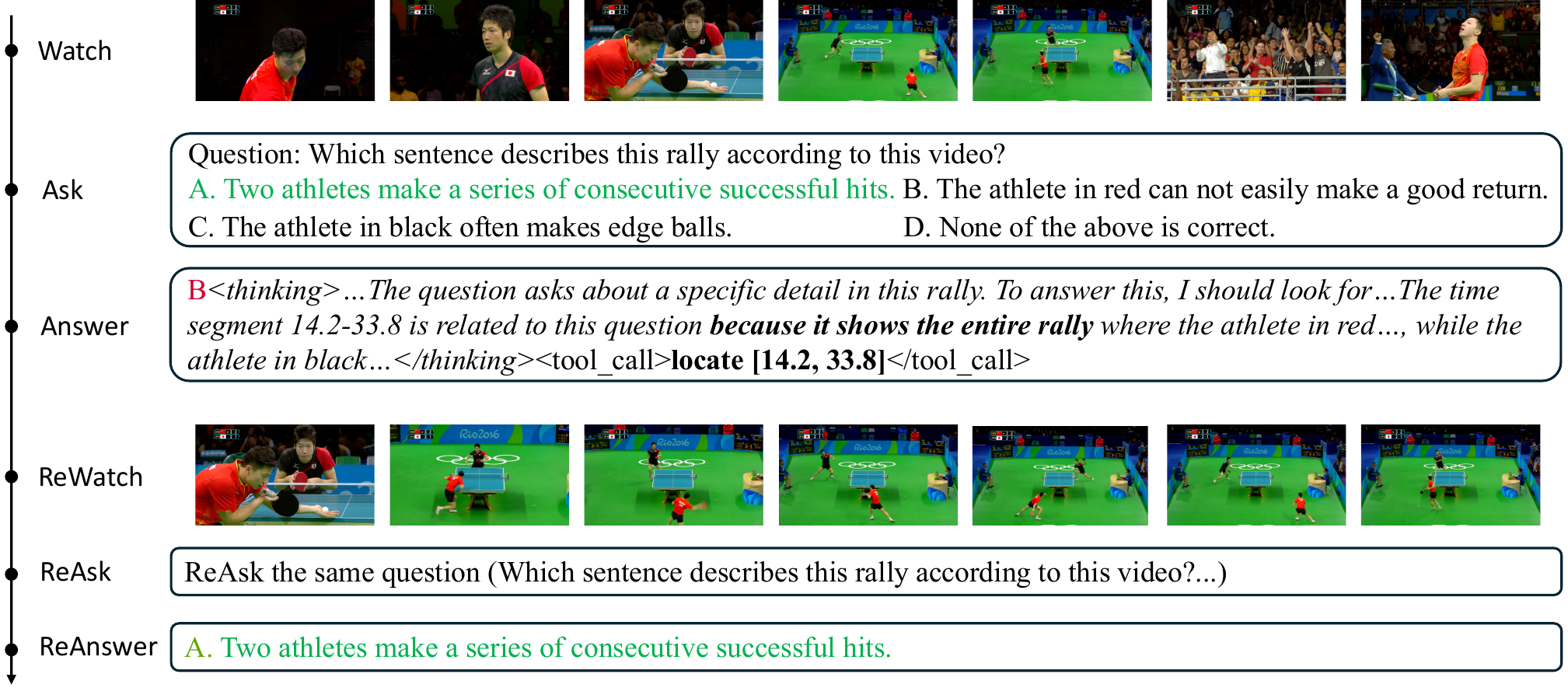}
    %\caption{The model workflow of video-SALMONN-R$^3$. Given a video and a user question, the model first \textbf{Watches} the full video at low resolution and low frame rate, \textbf{Asks} the query, and produces an initial \textbf{Answer}, followed by a reasoning process and a locate tool call that pinpoints the relevant temporal segment. Conditioned on this grounded evidence, together with the first-pass video and Ask–Answer trace retained as context, the model \textbf{Re-Watch}es the localized clip at high resolution and high frame rate alongside the original low-fidelity full video, \textbf{Re-Ask}s the same question, and \textbf{Re-Answer}s, either keeping or revising the initial response. By jointly leveraging these two complementary streams (global low-fidelity context and local high-fidelity detail), the paradigm forms the backbone of video-SALMONN-R$^3$ for fine-grained video understanding.}
    \caption{Overview of the video-SALMONN-R$^3$ workflow. Given a video and a question, the model first \textbf{watches} the full video at low temporal and spatial fidelity, \textbf{asks} the query, and produces an initial \textbf{answer}, followed by a reasoning trace and a temporal localization tool call. Conditioned on the localized evidence, together with the first-pass context, the model \textbf{re-watches} the selected segment at higher fidelity, \textbf{re-asks} the question, and \textbf{re-answers} by either retaining or refining the initial answer. By combining global low-fidelity context with local high-fidelity detail, this two-pass design enables fine-grained video understanding.}
    \vspace*{-0.5cm}
    \label{fig:overall}
\end{figure}

%The overall workflow of video-SALMONN-R$^3$ is demonstrated in Fig.~\ref{fig:overall}. Following video-SALMONN 2 \cite{tang2025video}, the audio and visual tokens of the video are interleaved to obtain the video representation $V=\operatorname{Interleave}(V_{\text{video}}, V_{\text{audio}})$, which, together with the textual question input $Q$, is fed into the LLM to produce the output $O$. A single forward pass can be represented with Equ.~(\ref{equ:eq1}).

The workflow of video-SALMONN-R$^3$ is shown in Fig.~\ref{fig:overall}. %Following video-SALMONN 2 \cite{tang2025video}, 
We interleave visual and audio tokens produced by relevant encoders to form a unified video representation $\mathbf{V}=\operatorname{Interleave}(\mathbf{V}_{\text{video}}, \mathbf{V}_{\text{audio}})$, which, together with the question $\mathbf{Q}$, is fed into the LLM to produce an output $\mathbf{O}$:
\begin{equation}\label{equ:eq1}
    \mathbf{O}=\operatorname{LLM}(\textbf{V},\textbf{Q}).
\end{equation}

\textbf{First pass (watch).} The model processes the full video at reduced frame rate and spatial resolution to obtain a coarse global representation $\mathbf{V}^{(1)} = \operatorname{Interleave}(\mathbf{V}^{(1)}_{\text{video}}, \mathbf{V}^{(1)}_{\text{audio}})$. Conditioned on $\mathbf{V}^{(1)}$ and $\mathbf{Q}$, the LLM produces a structured output $\mathbf{O}^{(1)}$ comprising three components: an initial answer $\mathbf{A}^{(1)}$, a short reasoning trace $\mathbf{R}^{(1)}$, and a temporal localization prediction $\mathbf{T}=[t_\text{start}, t_\text{end}]$ identifying the segment most relevant to the query:
\begin{equation}\label{equ:eq2}
\mathbf{O}^{(1)} = (\mathbf{A}^{(1)}, \mathbf{R}^{(1)}, \mathbf{T}) = \operatorname{LLM}(\mathbf{V}^{(1)}, \mathbf{Q}).
\end{equation}
Here, $\mathbf{A}^{(1)}$ leverages the well-aligned prior of the instruction-tuned base model, while $\mathbf{T}$ determines where higher-fidelity processing will be applied in the next pass.

%In the first pass, the model watches the full video at a low frame rate and low spatial resolution to obtain a coarse global representation $V^{(1)} = \operatorname{Interleave}(V^{(1)}_{\text{video}}, V^{(1)}_{\text{audio}})$. Conditioned on $V^{(1)}$ and the user question $Q$, the LLM is prompted to produce a structured output $O^{(1)}$ that contains three parts: an initial answer $A^{(1)}$ drawn from the well-aligned prior of the instruction-tuned base model, a short reasoning trace $R^{(1)}$, and a temporal localization tool call $T = [t_s, t_e]$ that pinpoints the segment most relevant to $Q$, as shown in Equ.~(\ref{equ:eq2}).
%\begin{equation}\label{equ:eq2}
%    O^{(1)} = (A^{(1)}, R^{(1)}, T) = \operatorname{LLM}(V^{(1)}, Q).
%\end{equation}

%The initial answer $A^{(1)}$ anchors the subsequent refinement onto existing competence, while the predicted interval $T$ determines where the model should re-watch at higher fidelity in the next round.

\textbf{Second pass (re-watch).} Guided by $\textbf{T}$, the model revisits the localized segment at higher temporal and spatial fidelity, yielding a fine-grained representation $\mathbf{V}^{(2)} = \operatorname{Interleave}(\mathbf{V}^{(2)}_{\text{video}}, \textbf{V}^{(2)}_{\text{audio}})$ sampled from $[t_\text{start}, t_\text{end}]$. To integrate global context and local detail, we retain the first-pass input $\mathbf{V}^{(1)}$ and the interaction trace $(\textbf{Q}, \textbf{O}^{(1)})$. The question is then re-injected as $\textbf{Q}'$ after $\textbf{V}^{(2)}$, enabling direct attention between the query and the newly observed frames under causal attention. The refined answer is produced as:
\begin{equation}\label{equ:eq3}
\textbf{O}^{(2)} = \textbf{A}^{(2)} = \operatorname{LLM}\big(\textbf{V}^{(1)}, \textbf{Q}, \textbf{O}^{(1)}, \textbf{V}^{(2)}, \textbf{Q}'\big).
\end{equation}

The final prediction $\mathbf{A}^{(2)}$ either retains $\mathbf{A}^{(1)}$ when the initial evidence suffices or revises it when higher-fidelity inspection reveals additional details. This two-pass watch–re-watch loop forms the core computational unit of video-SALMONN-R$^3$ and serves as the trajectory for subsequent RL.

\vspace*{-0.1cm}
\subsection{Training Procedure}
\vspace*{-0.1cm}

Training proceeds in three stages: \textbf{1)} {audio alignment}, \textbf{2)} {audio-visual caption SFT}, and \textbf{3)} {end-to-end RL}. The first two stages establish a strong instruction-tuned multimodal base model, while the final stage injects re-watch behavior without CoT-based cold-start. %Note in this paper audio is included in video. 
Unless otherwise specified, we treat audio as an integral component of the video modality throughout this paper.

%The training of video-SALMONN-R$^3$ proceeds in three stages. We first perform \textbf{audio alignment} \cite{tang2024salmonn}, training only the audio projector to map acoustic features into the LLM's representation space. We then conduct \textbf{audio-visual caption SFT} \cite{tang2025video} on densely annotated audio-visual captions, yielding an instruction-tuned base model with well-aligned audio-visual competence. Finally, we apply \textbf{end-to-end RL} on top of this baseline, optimizing the full trajectory with DAPO \cite{yudapo} to inject temporal localization and re-watch behaviors without any CoT cold-start.

\textbf{Audio alignment.} We first augment a visual-only LLM with an audio branch by training an audio projector that maps acoustic features into the LLM representation space \cite{tang2024salmonn}. During this stage, only the projector is updated, while the audio encoder, visual encoder, and LLM backbone remain frozen. Training uses large-scale speech recognition and audio captioning data under a standard cross-entropy objective. This step establishes a stable interface between audio signals and the LLM without disturbing existing visual capabilities.

\textbf{Video caption SFT.} Next, we jointly model audio-visual inputs by fine-tuning on densely annotated video captions \cite{tang2025video}. The model is trained to generate unified textual descriptions of video content from interleaved audio and visual tokens. To enable efficient adaptation while preserving pretrained knowledge, we employ low-rank adaptation (LoRA) \cite{hu2022lora} on the LLM backbone and jointly optimize it with the modality projectors, while keeping the audio and visual encoders frozen. This stage produces a strong instruction-tuned base video-LLM.

%\textbf{End-to-end RL} directly injects the re-watch capability on top of the instruction-tuned base model without any CoT cold-start SFT, which is costly to curate and tends to drag the aligned baseline toward the annotation pipeline's biases, eroding existing competence. Instead, a carefully designed system prompt specifies the protocol and the \texttt{<thinking>}/\texttt{<tool\_call>} schema, giving the model a minimal format prior and letting it explore full trajectories under rule-based rewards so that localization and re-watch emerge as refinements rather than overwrites. Crucially, by the re-answer design, everything emitted up to $A^{(1)}$ mirrors the baseline conditioned on $(V^{(1)}, Q)$, with no extra reasoning tokens, tool calls, or scaffolds inserted beforehand, so the first-pass answer is drawn from the baseline's original distribution and its competence is preserved by construction. See Appendix~\ref{app:prompt} for more training details in the RL stage.

\textbf{End-to-end RL.} Finally, we inject re-watch capability via RL, without relying on CoT cold-start SFT. Instead of constructing large-scale annotated reasoning trajectories, we define a lightweight system prompt specifying the interaction protocol and the \texttt{<thinking>} / \texttt{<tool\_call>} format, and optimize full trajectories using DAPO \cite{yudapo} with rule-based rewards.

Crucially, under the \textbf{re-answer} design, all tokens generated up to $\mathbf{A}^{(1)}$ follow the original base video-LLM distribution conditioned on $(\mathbf{V}^{(1)}, \textbf{Q})$, without introducing additional reasoning scaffolds or tool interactions beforehand. This ensures that the first-pass behavior remains aligned with the pretrained model, while the second-pass refinement introduces localization as an additive capability rather than a disruptive modification. Additional implementation details are provided in Appendix~\ref{app:prompt}.

\textbf{Reward design.} The trajectory reward is defined as a weighted sum of five rule-based components, each taking values in $\{0,1\}$:
\begin{itemize}[itemsep=1pt, leftmargin=*]
\item $r_{\text{acc1}}$, $r_{\text{acc2}}$: are accuracy rewards indicating whether the initial answer $\textbf{A}^{(1)}$ and the refined answer $\textbf{A}^{(2)}$ match the ground-truth option, respectively, ensuring correctness across both passes.
%\item $r_{\text{fmt1}}$: is a format reward indicating whether the first pass follows ``answer–think–locate'', ensuring that the tool call is well-formed and parseable; if this condition is not met, the re-watch stage is skipped.
\item $r_{\text{fmt1}}$: is a format reward indicating whether the first pass follows the ``answer--think--locate'' schema with a valid parseable tool call; otherwise, re-watch is skipped.
\item $r_{\text{fmt2}}$: is a format reward indicating whether the second pass contains no meta-tokens (\textit{e.g.}, $<$\texttt{thinking}$>$ for thinking and $<$\texttt{tool\_call}$>$ for tool calling), enforcing a clean final answer.
\item $r_{\text{rev}}$: is a revise reward that equals $0$ only when $\textbf{A}^{(1)}$ is incorrect and $\textbf{A}^{(2)}$ simply repeats it, and $1$ otherwise, encouraging the model to revise incorrect initial predictions.
\end{itemize}
The total reward for the $i\,$th trajectory $\textbf{o}_i$ is computed as
\begin{equation}
%\mathcal{R}(\textbf{o}_i) = \sum\nolimits_{k \in \mathcal{K}} \lambda_k, r_k(\textbf{o}_i),
\mathcal{R}(\textbf{o}_i) = \lambda_{\text{acc1}} r_{\text{acc1}}(\textbf{o}_i)+\lambda_{\text{acc2}} r_{\text{acc2}}(\textbf{o}_i)+\lambda_{\text{fmt1}} r_{\text{fmt1}}(\textbf{o}_i)+\lambda_{\text{fmt2}} r_{\text{fmt2}}(\textbf{o}_i)+\lambda_{\text{rev}} r_{\text{rev}}(\textbf{o}_i),
\end{equation}
where the ${\lambda}$s are coefficients that balance the contributions of each component. The group-normalized advantage for each token is then defined as
%where ${\lambda_k}$ are coefficients that balance the contributions of each component. The group-normalized advantage for each token is then defined as
\begin{equation}
\label{advantage}
\hat{\mathcal{A}}_{i,t} = \frac{\mathcal{R}(\mathbf{o}_i) - \operatorname{mean}(\mathcal R)}{\operatorname{std}(\mathcal R) + \epsilon},
\end{equation}
where $\epsilon$ is a small constant for numerical stability, and $\operatorname{mean}(\cdot)$ and $\operatorname{std}(\cdot)$ denote the mean and standard deviation. Collectively, these reward signals capture answer correctness, structural validity, and the desired re-watch behavior (\textit{i.e.}, revising the initial answer after higher-fidelity inspection) thereby guiding the model toward effective re-watch reasoning without requiring CoT supervision.

\textbf{RL objective.}
We adopt a fully on-policy variant of DAPO \cite{yudapo} as the training objective. For each question $q$, a group of $G$ trajectories $\{\mathbf{o}_i\}_{i=1}^{G}$ is sampled from the current policy $\pi_\theta$. The policy is updated using the token-level loss $\mathcal{J}$:
\begin{equation}
\mathcal{J}(\theta) = -\mathbb{E}_{{q},{\mathbf{o}_i}}\left[
\frac{1}{\sum_{i=1}^{G}|\mathbf{o}_i|}\sum_{i=1}^{G}\sum_{t=1}^{|\mathbf{o}_i|}
\log \pi_\theta(\mathbf{o}_{i,t}\mid q, \mathbf{o}_{i,<t})\cdot\hat{\mathcal A}_{i,t}
\right],
\end{equation}
where $\hat{\mathcal A}_{i,t}$ is the group-normalized advantage derived from the trajectory rewards in Eqn.~\eqref{advantage}.

\begin{figure}[ht]
\centering
\includegraphics[width=0.9\linewidth]{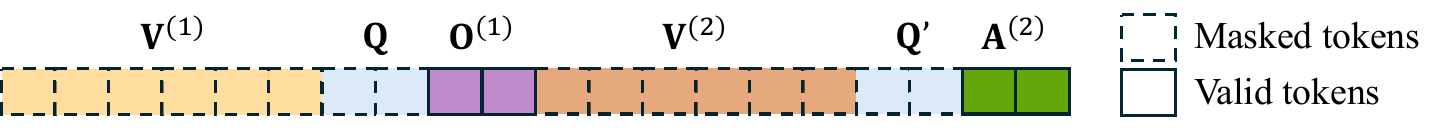}
\caption{Illustration of token selection for loss computation. Only valid tokens contribute to gradient updates.}
\label{fig:mask}
\end{figure}

\textbf{Training efficiency.}
For efficiency, we do not compute rewards or losses separately for each pass. Instead, we roll out the entire trajectory and optimize the loss in a single forward–backward pass over the full sequence. The loss is applied only to tokens generated by the model, including the first-pass output $\mathbf{O}^{(1)}$ (with reasoning $\mathbf{R}^{(1)}$, tool call $\mathbf{T}$, and initial answer $\mathbf{A}^{(1)}$) and the second-pass refined answer $\mathbf{A}^{(2)}$. All other tokens are masked out, as illustrated in Fig.~\ref{fig:mask}.

\vspace*{-0.1cm}
\section{Experiment Setup}\label{sec:set}
\vspace*{-0.1cm}

\subsection{Model Specifications}
\vspace*{-0.1cm}

%Built upon the Qwen3-VL 8B backbone \cite{bai2025qwen3}, video-SALMONN-R$^3$ inherits the audio-processing design of video-SALMONN-2 \citep{tang2025video}: audio features are extracted by the Whisper-Large-v3 encoder \citep{radford2023robust} and subsequently aligned through a window-level Q-Former with a 0.5-second window. 

Built upon a textual LLM backbone of Qwen3-VL with 8 billion (B) parameters \cite{bai2025qwen3}, the audio features of video-SALMONN-R$^3$ are extracted by the Whisper-Large-v3 encoder \citep{radford2023robust} and subsequently aligned through a window-level Q-Former with a 0.5-second window. 

In the first pass, visual frames are sampled at 10 frames per second (FPS) with a maximum of 768 frames. Videos longer than 76.8 seconds are uniformly sub-sampled to meet this budget, and each frame is capped at 44{,}100 pixels to maintain a broad but low-fidelity global view under a constrained token budget.
In the second pass, the localized segment is also sampled at 10 FPS, but with a stricter limit of 128 frames; segments longer than 12.8 seconds are uniformly down-sampled. In contrast to the first pass, each frame is allowed up to 176{,}400 pixels to preserve fine-grained spatial details within the selected interval.
Although this configuration can theoretically produce up to approximately 27{,}000 visual tokens across both passes, the actual count is further regulated by the visual pre-processor’s smart resizing policy, which keeps the total number of visual tokens around 23{,}000 per trajectory in practice.

%In the first pass, input visual frames are sampled at a frame rate of 10 frames per second (FPS) with an upper bound of 768 frames: videos longer than 76.8 sec. are uniformly sub-sampled down to this budget, and each frame is capped at 44{,}100 pixels to preserve a broad but low-fidelity global view within a limited token budget. In the second pass, the localized clip is likewise sampled at 10 FPS but with a tighter cap of 128 frames, so segments exceeding 12.8 sec. are uniformly down-sampled, while each frame is allowed up to 176{,}400 pixels to retain fine spatial detail within the grounded interval. Although this configuration can in principle yield up to roughly 27{,}000 visual tokens across the two passes, the actual token count is further constrained by the smart-resize policy of the visual pre-processor, which in practice keeps the total number of visual tokens at around 23{,}000 per trajectory.

\vspace*{-0.1cm}
\subsection{Training Specifications}
\vspace*{-0.1cm}

For the audio alignment stage, we train only the audio aligner while keeping the audio encoder, the visual branch, and the LLM backbone frozen, using a combination of LibriSpeech 960-hour \citep{panayotov2015librispeech} and CommonVoice \citep{ardila2020common} for automatic speech recognition, together with WavCaps \citep{mei2024wavcaps} and AudioCaps \citep{kim2019audiocaps} for audio captioning, so as to establish a reliable interface from acoustic features into the representation space of the LLM. For the audio-visual caption SFT stage, we fine-tune the model on audio-visual captions derived from LLaVA-Video-178k \citep{zhang2024video} and re-annotated with Gemini 2.5 Pro \citep{comanici2025gemini}, which jointly describe the visual and acoustic content within a unified temporal narrative and yield a strong instruction-tuned audio-visual base model. 

%For the end-to-end RL stage, we use CinePile \citep{rawal2024cinepile} and CG-Bench \citep{chen2024cg} as the short- and long-video training sources, respectively, covering a broad spectrum of temporal scales. To improve the difficulty distribution, we run 8 rollouts with the audio-visual caption SFT model over the full training set and discard samples answered correctly or incorrectly in all 8 attempts, retaining about 110k moderately difficult samples. We use a learning rate of $5\times10^{-6}$, a group size of 8, and reward coefficients $\lambda_\text{acc1}=0.9, \lambda_\text{acc2}=1.1, \lambda_\text{fmt1}=1, \lambda_\text{fmt2}=1, \lambda_\text{rev}=0.5$ to encourage stronger re-answers. The reward curves can be found in Appendix~\ref{app:curve}. More ablation on the RL setting can be found in Appendix~\ref{app:abl}.

For the end-to-end RL stage, we use CinePile \citep{rawal2024cinepile} and CG-Bench \citep{chen2024cg} as training sources for short- and long-form videos, respectively, covering a broad range of temporal scales. To obtain a more informative difficulty distribution, we perform 8 rollouts using the audio-visual caption SFT model over the entire training set and discard samples that are answered either correctly or incorrectly in all attempts, retaining approximately 110k moderately challenging samples.
We use a learning rate of $5\times10^{-6}$, a group size of 8, and reward coefficients $\lambda_\text{acc1}=0.9$, $\lambda_\text{acc2}=1.1$, $\lambda_\text{fmt1}=1$, $\lambda_\text{fmt2}=1$, and $\lambda_\text{rev}=0.5$, slightly emphasizing the refined answer during optimization. Reward curves are provided in Appendix~\ref{app:curve}, and additional ablations on the RL settings can be found in Appendix~\ref{app:abl}.

\vspace*{-0.1cm}
\subsection{Evaluation Specifications}
\vspace*{-0.1cm}

We evaluate video-SALMONN-R$^3$ on both short-to-medium and long video benchmarks to comprehensively assess the proposed re-watch paradigm across different temporal scales. Regarding short-to-medium videos, we adopt \textit{VideoHolmes} \cite{cheng2025video}, \textit{DailyOmni} \cite{zhou2025daily}, \textit{AVUT} \cite{yang2025acvubench}, and \textit{OmniVideoBench} \cite{li2025omnivideobench}, which collectively focus on fine-grained audio-visual reasoning within tight temporal windows, where correctly answering the question hinges on localizing and jointly interpreting subtle acoustic and visual cues.
Regarding long videos, we adopt \textit{VideoMME} \cite{fu2024video} and \textit{LVOmniBench} \cite{tao2026lvomnibench}, which collectively focus on long-range temporal understanding over videos spanning tens of minutes to hours, where the model must locate query-relevant segments scattered across lengthy content and reason over distant temporal dependencies.

\vspace*{-0.1cm}
\section{Experiments}
\vspace*{-0.1cm}
\subsection{Overall Results}
\vspace*{-0.1cm}
% \begin{table*}[!ht]
%     \centering
%     \small
%     \setlength{\tabcolsep}{4pt}
%     \renewcommand{\arraystretch}{1.15}
%     \caption{Overall comparison on short-to-medium benchmarks (VH: VideoHolmes, DO: DailyOmni, AVUT, OVB: OmniVideoBench) and long-form benchmarks (VMME: Video-MME, LVOB: LVOmniBench).}
%     \label{tab:overall}
%     \begin{tabular}{l|cccc|cc}
%         \toprule
%         \multirow{2}{*}{\textbf{Model}} & \multicolumn{4}{c|}{\textbf{Short-to-Medium}} & \multicolumn{2}{c}{\textbf{Long}} \\
%         \cmidrule(lr){2-5} \cmidrule(lr){6-7}
%          & VH & DO & AVUT & OVB & VMME & LVOB \\
%         \midrule
%         Qwen 2.5-Omni & 43.7 & 62.7 & 66.3 & 29.3 & 64.3 & 32.0 \\
%         video-SALMONN 2+ & 46.9 & 71.8 & 69.5 & 36.4 & 73.4 & 32.7 \\
%         Qwen 3-Omni & 54.1 & 69.8 & 72.0 & 38.4 & 70.5 & 35.8 \\
%         D-ORCA & 48.5 & 78.5 & 76.1 & - & 72.9 & - \\
%         \midrule
%         AV-Caption-SFT& 49.8 & 76.7 & 76.7 & 40.0 & 72.5 & 39.3 \\
%         QA-SFT & 53.4 & 77.9 & 74.1 & 40.8 & 72.9 & 40.6 \\
%         \rowcolor{gray!10}
%         \textbf{video-SALMONN-R$^3$ (ours)} & \textbf{54.6} & \textbf{78.7} & \textbf{77.5} & \textbf{43.9} & \textbf{76.3} & \textbf{42.9} \\
%         \bottomrule
%     \end{tabular}
% \end{table*}

\begin{table*}[!ht]
    \centering
    %\footnotesize
    \small
    \setlength{\tabcolsep}{2pt}
    \renewcommand{\arraystretch}{1.15}
    %\caption{Overall comparison on short-to-medium benchmarks and long-form benchmarks to the strongest audio-visual models with similar scales (7/8B Dense or 30BA3B MoE). video-SALMONN-R$^3$ consistently outperforms baselines on all benchmarks.}
    \caption{Overall comparison on short-to-medium and long-form benchmarks against SOTA audio-enabled video-LLMs of comparable scale (7B/8B dense or 30B A3B mixture-of-expert). video-SALMONN-R$^3$ consistently achieves superior performance across all benchmarks.}
    \label{tab:overall}
    \begin{tabular}{l|cccc|cc}
        \toprule
        \multirow{2}{*}{\textbf{Model}} & \multicolumn{4}{c|}{\textbf{Short-to-Medium}} & \multicolumn{2}{c}{\textbf{Long}} \\
        \cmidrule(lr){2-5} \cmidrule(lr){6-7}
         & VideoHolmes & DailyOmni & AVUT & OmniVideoBench & VideoMME & LVOmniBench \\
        \midrule
        Qwen 2.5-Omni \cite{xu2025qwen2} & 43.7 & 62.7 & 66.3 & 29.3 & 64.3 & 32.0 \\
        video-SALMONN 2+ \cite{tang2025video} & 46.9 & 71.8 & 69.5 & 36.4 & 73.4 & 32.7 \\
        Qwen 3-Omni \cite{xu2025qwen3} & 54.1 & 69.8 & 72.0 & 38.4 & 70.5 & 35.8 \\
        D-ORCA \cite{tang2026dorca} & 48.5 & 78.5 & 76.1 & - & 72.9 & - \\
        \midrule
        Qwen 3-VL & 46.6 & 60.1 & 61.4 & - & 71.4 & 35.6 \\
        AV-Caption-Base& 49.8 & 76.7 & 76.7 & 40.0 & 72.5 & 39.3 \\
        QA-SFT & 53.4 & 77.9 & 74.1 & 40.8 & 72.9 & 40.6 \\
        \rowcolor{gray!10}
        %\textbf{video-SALMONN-R$\mathbf{^3}$ (ours)} & \textbf{54.6} & \textbf{78.7} & \textbf{77.5} & \textbf{43.9} & \textbf{76.3} & \textbf{42.9} \\
        \textbf{video-SALMONN-R$\mathbf{^3}$} & \textbf{54.6} & \textbf{78.7} & \textbf{77.5} & \textbf{43.9} & \textbf{76.3} & \textbf{42.9} \\
        \bottomrule
    \end{tabular}
    \vspace*{-0.3cm}
\end{table*}

%Table~\ref{tab:overall} reports the overall comparison across 6 benchmarks spanning short-to-medium and long-form settings. Alongside frontier audio-visual LLMs, we include two in-house controls: AV-Caption-Base, sharing the same backbone and captioning pre-training as our model, and QA-SFT, trained on the exact same QA data as video-SALMONN-R$^3$ but with SFT instead of RL. This setup isolates the contribution of the RL-induced re-watch behavior from confounding factors such as training data or backbone capacity.

Table~\ref{tab:overall} presents the overall comparison across six benchmarks spanning short-to-medium and long-form video settings. In addition to frontier audio-visual LLMs, we include two in-house baselines: AV-Caption-Base, which shares the same backbone and caption pre-training as our model, and QA-SFT, which is trained on the same audio-visual question answering (QA) data as video-SALMONN-R$^3$ but uses SFT instead of RL. This setup isolates the effect of RL-induced re-watch behavior from confounding factors such as training data and model capacity.

%Across all temporal scales, video-SALMONN-R$^3$ consistently surpasses the strongest open audio-visual LLMs of comparable size. The gains span reasoning-heavy (VideoHolmes), perception-dense (AVUT), and cross-modal alignment (DailyOmni, OmniVideoBench) tasks, indicating that the re-watch policy generalizes rather than overfits to a single format. Notably, the advantage widens on long-form benchmarks (+3.4 on VideoMME and +7.1 on LVOmniBench over the best baseline), suggesting that selectively revisiting salient segments becomes increasingly valuable as video length grows and uniform sampling becomes lossy.

Across all temporal scales, video-SALMONN-R$^3$ consistently outperforms the previous SOTA open-source audio-enabled video-LLMs of comparable scale. The improvements span reasoning-oriented (VideoHolmes), perception-intensive (AVUT), and cross-modal alignment (DailyOmni, OmniVideoBench) benchmarks, suggesting that the learned re-watch policy generalizes across diverse task formats rather than specializing to a particular benchmark. Notably, the gains become larger on long-form benchmarks (+3.4 on VideoMME and +7.1 on LVOmniBench over the strongest baseline), indicating that selectively revisiting salient segments becomes increasingly beneficial as video duration grows and uniform sampling becomes less effective.

%The comparison against QA-SFT is particularly informative: despite identical QA supervision, video-SALMONN-R$^3$ yields consistent gains across all benchmarks, with the largest margins on long-form and reasoning-intensive tasks. This gap shows that re-watch behavior cannot be imitated purely from answer-level supervision. It emerges only under an outcome-driven RL objective. Taken together, these results demonstrate that re-watch acquired purely through RL yields tangible and scalable gains across temporal granularities and reasoning types, without sacrificing general video competence.

The comparison with QA-SFT is particularly revealing. Despite using identical QA supervision, video-SALMONN-R$^3$ consistently achieves stronger performance across all benchmarks, with the largest improvements appearing on long-form and reasoning-intensive tasks. This result suggests that re-watch behavior is difficult to acquire from answer-level supervision alone, and instead emerges more naturally under an outcome-driven RL objective. Overall, these results demonstrate that re-watch capability learned purely through RL provides scalable gains across temporal granularities and reasoning types, while preserving general video understanding performance.

\begin{wraptable}{r}{0.54\linewidth}
    \centering
    \small
    \vspace*{-0.3cm}
    \setlength{\tabcolsep}{4pt}
    \renewcommand{\arraystretch}{1.15}
    \caption{Comparison of Video-MME against existing localization-based video LLMs.}
    \label{tab:vmme_loc}
    \begin{tabular}{l|cccc}
        \toprule
        \textbf{Model} & \textbf{Short} & \textbf{Medium} & \textbf{Long} & \textbf{Avg} \\
        \midrule
        \multicolumn{5}{l}{\textit{Multi-agent localization}} \\
        \midrule
        VideoLucy   & 78.6 & 72.1 & 66.8 & 72.5 \\
        GCAgent     & 72.6 & 69.8 & \textbf{73.4} & 71.9 \\
        \midrule
        \multicolumn{5}{l}{\textit{Single-model localization}} \\
        \midrule
        VideoChat-R1.5 & - & - & - & 67.1 \\
        LOVE-R1        & 75.3 & 65.6 & 57.7 & 66.2 \\
        LongVT         & - & - & - & 67.0 \\
        VideoZoomer    & - & - & 55.8 & 65.2 \\
        \midrule
        \rowcolor{gray!10}
        \textbf{video-SALMONN-R$\mathbf{^3}$} & \textbf{83.2} & \textbf{78.6} & 67.2 & \textbf{76.3} \\
        \bottomrule
    \end{tabular}
    \vspace*{-0.3cm}
\end{wraptable}

Table~\ref{tab:vmme_loc} compares video-SALMONN-R$^3$ with prior localization-based approaches on Video-MME. Multi-agent methods (\textit{e.g.}, VideoLucy and GCAgent) combine powerful external text reasoners such as GPT-5.1 and DeepSeek-R1 with large collections of caption memories (\textit{e.g.}, over 120 captions per minute for a 2-hour video). These approaches perform especially well on the Long split (\textit{e.g.}, GCAgent achieves 73.4), surpassing all models in Table~\ref{tab:overall}. This highlights the advantage of aggregating evidence from dense captions with strong text-based reasoning. However, their gains on the Short and Medium splits are more limited, likely because the final reasoning stage relies on textual summaries instead of direct video inputs and is therefore constrained by caption quality.
Single-model localization methods address this limitation by integrating localization and reasoning within a unified architecture. However, they typically depend on costly CoT cold-start supervision, which can affect the alignment of the pretrained base model. Consequently, their overall performance generally remains in the 65--67 range. In contrast, video-SALMONN-R$^3$ achieves the best overall performance, suggesting that the combination of re-answer, re-ask, and cold-start-free RL provides a more balanced and efficient framework for fine-grained re-watch-based temporal localization.

\vspace*{-0.1cm}
\subsection{Analysis of Re-Watch}
\vspace*{-0.1cm}

\begin{table}[ht]
    \centering
    %\footnotesize
    \small
    \setlength{\tabcolsep}{4pt}
    \renewcommand{\arraystretch}{1.15}
    %\caption{Ablation isolating the effect of re-watch. All variants report the final answer $\mathbf{A}^{(2)}$, except QA-SFT, which is single-pass and reported as its sole output. \emph{Re-Answer only} feeds the same low-fidelity video to both passes; \emph{Uniform Re-Watch} uses the same total visual-token budget as the full model but samples uniformly over the whole video instead of zooming into $T$. The gap between these baselines and the full model quantifies the information genuinely contributed by temporally targeted re-watching.}
    \caption{Ablation study isolating the effect of re-watch. All variants report the final answer $\mathbf{A}^{(2)}$, except QA-SFT, which is single-pass and reported using its sole output. \emph{Re-Answer only} feeds the same low-fidelity video into both passes, while \emph{Uniform Re-Watch} uses the same total visual-token budget as the full model but samples uniformly across the entire video instead of revisiting the localized segment $\mathbf{T}$. The performance gap between these baselines and the full model reflects the contribution of temporally targeted re-watching.}
    \label{tab:rewatch_ablation}
    \begin{tabular}{l|cccc}
        \toprule
        \textbf{Variant} & VideoHolmes & DailyOmni & VideoMME & LVOmniBench \\
        \midrule
        QA-SFT          & 53.4 & 77.9 & 72.9 & 40.6 \\
        Re-Answer only                & 53.1 & 77.3 & 74.4 & 39.7 \\
        Uniform Re-Watch              & 52.6 & 78.4 & 73.9 & 41.0 \\
        \textbf{video-SALMONN-R$\mathbf{^3}$ (full)} & \textbf{54.6} & \textbf{78.7} & \textbf{76.3} & \textbf{42.9} \\
        \bottomrule
    \end{tabular}
    \vspace*{-0.5cm}
\end{table}

%To verify that the gains of video-SALMONN-R$^3$ stem from temporally targeted re-watching rather than from answering twice or simply consuming more visual tokens, we compare four variants that share the same backbone and training data and differ only in the second-pass visual context, with results reported in Table~\ref{tab:rewatch_ablation}. \emph{QA-SFT} watches the video once at low fidelity and serves as the single-pass reference. \emph{Re-Answer only} reanswers the question after thinking without additional video inputs and barely lifts performance over the single-pass reference, confirming that a second answering pass without new visual evidence brings no real gain. \emph{Uniform Re-Watch} matches the full model's sampling setting (lower frames but higher fidelity) but samples the second-pass frames uniformly over the whole video instead of zooming into $T$, and still trails the full model by a clear margin, ruling out the trivial explanation that more tokens alone suffice. \emph{video-SALMONN-R$^3$ (full)} concentrates the same budget on the predicted interval $T$ and pulls ahead of all baselines, with the widest margins on benchmarks that reward fine-grained (VideoHolmes, DailyOmni) or scattered long-range evidence (VideoMME, LVOmniBench), showing that the locate tool call genuinely routes the limited visual-token budget toward the relevant region. More cases can be found in Appendix~\ref{app:case}.

To verify that the gains of video-SALMONN-R$^3$ arise from temporally targeted re-watching rather than simply answering twice or consuming more visual tokens, we compare four variants that share the same backbone and training data, differing only in the second-pass visual context. Results are reported in Table~\ref{tab:rewatch_ablation}.
\emph{QA-SFT} processes the video only once at low fidelity and serves as the single-pass baseline. \emph{Re-Answer only} performs a second answering pass after reasoning but without additional video inputs, yielding only marginal improvement over the single-pass baseline and suggesting that re-answering alone provides limited benefit without new visual evidence. \emph{Uniform Re-Watch} matches the full model in token budget and second-pass sampling configuration (fewer frames but higher fidelity), but uniformly samples frames across the entire video instead of revisiting the localized segment $\mathbf{T}$. Despite using the same number of tokens, it still underperforms the full model by a clear margin, indicating that the gains cannot be explained solely by increased visual input.
In contrast, \emph{video-SALMONN-R$^3$ (full)} concentrates the same visual-token budget on the predicted interval $\mathbf{T}$ and consistently outperforms all baselines. Additional qualitative examples are provided in Appendix~\ref{app:case}.

\vspace*{-0.1cm}
\subsection{Analysis of Re-Ask}
\vspace*{-0.1cm}

%To disentangle the effect of re-ask on the two passes of answers, we compare configurations that differ in whether the question is re-posed after the second-pass video, reporting the accuracy of $\mathbf{A}^{(1)}$ and $\mathbf{A}^{(2)}$ together with the average attention mass of all layers the re-answer token places on three anchors: the first-pass question ($\mathbf{a}_\mathbf{Q}$), the committed first-pass answer ($\mathbf{a}_{\mathbf{A}^{(1)}}$), and the re-posed question ($\mathbf{a}_{\mathbf{Q}'}$), averaged over 50 randomly sampled test examples. In the course of this analysis, we further find that re-ask alone is not enough: the revise reward $r_{\text{rev}}$ turns out to be an equally indispensable ingredient, so we additionally include a setting that enables $\mathbf{Q}'$ but disables $r_{\text{rev}}$ to expose their interplay.

To isolate the effect of re-ask on the two answering stages, we compare configurations that differ in whether the question is re-posed after the second-pass video. We report the accuracies of $\mathbf{A}^{(1)}$ and $\mathbf{A}^{(2)}$, together with the average attention mass assigned by the re-answer token across all layers to three anchors: the first-pass question ($\mathbf{a}
_{\mathbf{Q}}$), the committed first-pass answer ($\mathbf{a}_{\mathbf{A}^{(1)}}$), and the re-posed question ($\mathbf{a}_{\mathbf{Q}'}$). The statistics are averaged over 50 randomly sampled test examples.
During this analysis, we further observe that re-ask alone is insufficient. The revise reward $r_{\text{rev}}$ is also critical for encouraging effective refinement after re-watching. To study their interaction, we additionally include a setting that enables $\mathbf{Q}'$ while disabling $r_{\text{rev}}$.

\begin{table}[ht]
    \centering
    %\footnotesize
    \small
    \vspace*{-0.5cm}
    \setlength{\tabcolsep}{4pt}
    \renewcommand{\arraystretch}{1.15}
    %\caption{Accuracy and re-answer attention mass under three RL settings. Re-ask is the primary object of study; the third row is added to reveal that the revise reward $r_{\text{rev}}$ is equally indispensable: without it, $\mathbf{Q}'$ alone fails to activate a genuine refinement.}
    \caption{Accuracy and re-answer attention mass under three RL settings. Re-ask is the primary factor under investigation, while the third setting shows that the revise reward $r_{\text{rev}}$ is also essential: without it, the re-posed question $\mathbf{Q}'$ alone is insufficient to induce effective answer refinement.}
    \label{tab:reask_ablation}
    \begin{tabular}{l|l|cccc|ccc}
        \toprule
        \multirow{2}{*}{Setting} & \multirow{2}{*}{Pass} & \multicolumn{4}{c|}{Accuracy} & \multicolumn{3}{c}{Attention Mass} \\
        \cmidrule(lr){3-6} \cmidrule(lr){7-9}
         & & VideoHolmes & DailyOmni & VideoMME & LVOmniBench & $\mathbf{a}_\mathbf{Q}$ & $\mathbf{a}_{\mathbf{A}^{(1)}}$ & $\mathbf{a}_{\mathbf{Q}'}$ \\
        \midrule
        \multirow{2}{*}{w/o $\mathbf{Q}'$, w/ $r_{\text{rev}}$} & $\mathbf{A}^{(1)}$ & 52.6 & 77.3 & 71.9 & 38.5 & \multirow{2}{*}{0.09} & \multirow{2}{*}{0.06} & \multirow{2}{*}{-} \\
         & $\mathbf{A}^{(2)}$ & 52.7 & 77.3 & 71.9 & 38.5 \\
        \midrule
        \multirow{2}{*}{w/ $\mathbf{Q}'$, w/o $r_{\text{rev}}$} & $\mathbf{A}^{(1)}$ & 53.0 & 78.0 & 73.6 & 42.6 & \multirow{2}{*}{0.10} & \multirow{2}{*}{0.08} & \multirow{2}{*}{0.04} \\
         & $\mathbf{A}^{(2)}$ & 53.0 & 77.9 & 73.6 & 42.6 \\
        \midrule
        \multirow{2}{*}{w/ $\mathbf{Q}'$, w/ $r_{\text{rev}}$}  & $\mathbf{A}^{(1)}$ & 53.1 & 76.3 & 73.6 & 40.8 & \multirow{2}{*}{0.03} & \multirow{2}{*}{0.02} & \multirow{2}{*}{0.24} \\
        & $\mathbf{A}^{(2)}$ & \textbf{54.6} & \textbf{78.7} & \textbf{76.3} & \textbf{42.9} \\
        \bottomrule
    \end{tabular}
    
\end{table}

%Focusing first on re-ask, Table~\ref{tab:reask_ablation} shows that without $Q'$, $A^{(2)}$ is essentially identical to $A^{(1)}$ on all four benchmarks, so the second pass degenerates into an inert copy of the first. The attention numbers explain why: the re-answer tokens place comparable mass on $a_Q$ and $a_{A^{(1)}}$, so $A^{(2)}$ is strongly anchored on the already-committed answer and has no freshly-posed query adjacent to the re-watched frames to latch onto. With $Q'$ enabled (full setting), this balance is sharply reversed, and $A^{(2)}$ lifts clearly above $A^{(1)}$, confirming that re-ask is what actually turns the second pass into a genuine refinement step.

Focusing first on re-ask, Table~\ref{tab:reask_ablation} shows that without $\mathbf{Q}'$, the refined answer $\mathbf{A}^{(2)}$ remains nearly identical to $\mathbf{A}^{(1)}$ across all four benchmarks, causing the second pass to collapse into a passive repetition of the first. The attention statistics help explain this behavior: the re-answer token assigns comparable attention mass to $\mathbf{a}_\mathbf{Q}$ and $\mathbf{a}_{\mathbf{A}^{(1)}}$, indicating that $\mathbf{A}^{(2)}$ is strongly anchored to the committed first-pass answer, while lacking a newly posed query that can directly interact with the re-watched frames. Once $\mathbf{Q}'$ is introduced (full setting), this balance shifts substantially, and $\mathbf{A}^{(2)}$ consistently improves over $\mathbf{A}^{(1)}$, suggesting that re-ask is critical for making the second pass effective.
%suggesting that re-ask is critical for turning the second pass into an effective refinement stage.

%At the same time, we find that the revise reward $r_{\text{rev}}$ is an equally indispensable ingredient. In the middle row, adding $Q'$ but dropping $r_{\text{rev}}$ leaves $A^{(2)}$ virtually unrevised from $A^{(1)}$, and the attention on $Q'$ stays at a mere $0.04$. The reason is that once $A^{(1)}$ already collects most of the accuracy reward, the policy has no gradient signal to revise a confident first guess and learns to simply echo it, leaving the structurally available $Q'$ unused. Only when $Q'$ and $r_{\text{rev}}$ are enabled together does the attention actually flow through $Q'$ and the re-watched frames, delivering the observed accuracy gain. Re-ask thus opens a causal pathway from re-watched evidence to the refined answer, while $r_{\text{rev}}$ supplies the optimization pressure that ensures this pathway is actually used.

At the same time, we find that the revis reward $r_{\text{rev}}$ is equally important. In the middle-row setting in Table~\ref{tab:reask_ablation}, enabling $\mathbf{Q}'$ alone while removing $r_{\text{rev}}$ leaves $\mathbf{A}^{(2)}$ almost unchanged from $\mathbf{A}^{(1)}$, and the attention mass on $\mathbf{Q}'$ remains very small ($0.04$). Intuitively, once $\mathbf{A}^{(1)}$ already receives most of the accuracy reward, the policy lacks sufficient incentive to revise a confident initial prediction and instead learns to simply repeat it, leaving the structurally available $\mathbf{Q}'$ underutilized. Only when both $\mathbf{Q}'$ and $r_{\text{rev}}$ are enabled does attention meaningfully flow through the re-posed query and the re-watched frames, leading to clear accuracy improvements. These results suggest that re-ask establishes the causal pathway from re-watched evidence to the refined answer, while $r_{\text{rev}}$ provides the optimization pressure necessary to make effective use of this pathway.

\vspace*{-0.1cm}
\subsection{Analysis of Re-Answer \& RL Only Training}
\vspace*{-0.1cm}

%To isolate the contribution of our cold-start-free RL and the re-answer mechanism, we compare 3 more new variants against our full model. \emph{$A^{(2)}$ only} strips the re-answer mechanism, producing $A^{(2)}$ directly without first committing to $A^{(1)}$, which is trained with only $r_\text{fmt1}$, $r_\text{fmt2}$ and $r_\text{acc2}$. \emph{Localization SFT, w/o reasoning} and \emph{Localization SFT, w/ reasoning} follow the conventional cold-start recipe on CG-Bench localization annotations, with the latter additionally training CoT traces annotated by Gemini 2.5 Pro.

To isolate the contributions of the cold-start-free RL strategy and the re-answer mechanism, we compare three additional variants against the full model. $\mathbf{A}^{(2)}$ \emph{only} removes the re-answer design and directly generates $\mathbf{A}^{(2)}$ without first committing to $\mathbf{A}^{(1)}$; this variant is trained using only $r_{\text{fmt1}}$, $r_{\text{fmt2}}$, and $r_{\text{acc2}}$.
We further include two localization-SFT baselines following the conventional cold-start pipeline on CG-Bench localization annotations. \emph{Localization SFT, w/o reasoning} is trained only on localization supervision, while \emph{Localization SFT, w/ reasoning} additionally incorporates CoT reasoning traces annotated by Gemini 2.5 Pro.

\begin{table}[ht]
    \centering
    %\footnotesize
    \vspace*{-0.1cm}
    \small
    \setlength{\tabcolsep}{4pt}
    \renewcommand{\arraystretch}{1.15}
    %\caption{Ablation on the training recipe. All variants report the final answer produced by the model. Every variant that injects localization with other methods cannot significantly outperform the QA-SFT baseline, while our full recipe obtains substantial gains.}
    \caption{Ablation study on the training recipe. All variants report the model’s final answer. Variants that introduce localization through alternative training strategies achieve only marginal improvements over the QA-SFT baseline, whereas the full video-SALMONN-R$^3$ recipe delivers substantial gains.}
    \label{tab:reanswer_ablation}
    \begin{tabular}{l|cccc}
        \toprule
        \textbf{Variant} & VideoHolmes & DailyOmni & VideoMME & LVOmniBench \\
        \midrule
        QA-SFT               & 53.4 & 77.9 & 72.9 & 40.6 \\
        $\mathbf{A}^{(2)}$ only                   & 53.9 & 77.4 & 73.1 & 39.6 \\
        Localization SFT, w/o reasoning  & 53.2 & 77.3 & 72.9 & 40.4 \\
        Localization SFT, w/ reasoning   & 40.3 & 70.1 & 64.2 & 31.7 \\
        \textbf{video-SALMONN-R$\mathbf{^3}$}(full)  & \textbf{54.6} & \textbf{78.7} & \textbf{76.3} & \textbf{42.9} \\
        \bottomrule
    \end{tabular}
\end{table}

%Table~\ref{tab:reanswer_ablation} shows that every variant injecting localization without our re-answer scaffold fails to surpass the QA-SFT baseline: \emph{$A^{(2)}$ only} performs on par with the baseline, since without an anchored $A^{(1)}$ drawn from the well-aligned prior, the extra re-watch pass fails to translate into a stable end-task gain; \emph{Localization SFT, w/o reasoning} likewise leaves QA accuracy essentially flat; and \emph{Localization SFT, w/ reasoning}, the conventional CoT cold-start recipe, is the worst of all, as distilled reasoning traces carry biases alien to the instruction-tuned base model and erode its competence. Our full recipe instead lifts accuracy clearly above the baseline, confirming that re-answer acts as a competence-preserving anchor upon which rewatch-plus-reask converts localization into genuine end-task gains.

Table~\ref{tab:reanswer_ablation} shows that variants introducing localization without our re-answer scaffold fail to meaningfully outperform the QA-SFT baseline. $\mathbf{A}^{(2)}$ \emph{only} performs comparably to the baseline, suggesting that without an anchored first-pass answer $\mathbf{A}^{(1)}$ drawn from the well-aligned prior, the additional re-watch stage does not translate into stable downstream gains. \emph{Localization SFT, w/o reasoning} similarly yields little improvement in QA accuracy. Meanwhile, \emph{Localization SFT, w/ reasoning}, which follows the conventional CoT cold-start pipeline, performs worst among all variants, suggesting that externally distilled reasoning traces may introduce biases inconsistent with the instruction-tuned base model and negatively affect its existing capabilities.
In contrast, the full video-SALMONN-R$^3$ recipe achieves clear improvements over the baseline, indicating that re-answer serves as a competence-preserving anchor through which re-watch and re-ask can effectively translate temporal localization into downstream performance gains.

\vspace*{-0.1cm}
\section{Limitations}\label{sec:limit}
\vspace*{-0.1cm}

Our current design leaves three aspects open for future work. First, since our cold-start-free RL recipe places no explicit supervision on the \texttt{<thinking>} trace, the reasoning text and the predicted temporal interval occasionally lack a clear, interpretable correspondence. Second, both training and evaluation are restricted to multiple-choice QA, where rule-based rewards are easy to derive; extending the recipe to open-ended QA would require more sophisticated reward modeling such as LLM-as-a-judge. Third, the workflow performs temporal localization only once per question, which may be insufficient for videos whose query-relevant evidence is scattered across multiple distant segments and calls for iterative, multi-pass localization.

\vspace*{-0.1cm}
\section{Conclusion}
\vspace*{-0.1cm}

%We presented video-SALMONN-R$^3$, the first video LLM that acquires a \textbf{re-watch} temporal-localization capability purely through reinforcement learning on an instruction-tuned base model, entirely free of the costly chain-of-thought cold-start annotations that prior localization-based methods fundamentally depend on. To make this cold-start-free recipe viable, a \textbf{re-answer} mechanism anchors the newly acquired re-watch skill onto the baseline's well-aligned prior instead of overwriting it, while a \textbf{re-ask} mechanism remedies a causal-attention limitation so that the refined answer genuinely attends to the re-watched evidence. By first localizing the query-relevant segment and then re-watching it at high fidelity, the model routes its limited visual-token budget to where it matters most. Experiments across benchmarks show consistent gains over strong audio-visual baselines and prior localization-based video LLMs, and our ablations attribute these gains to targeted re-watching together with the re-ask and re-answer designs rather than to extra tokens or simply an additional answering pass. We hope this offers a simple and reproducible route toward efficient, fine-grained video understanding.

We presented video-SALMONN-R$^3$, the first video-LLM that acquires a \textbf{re-watch} temporal localization capability purely through RL on an instruction-tuned base model, without relying on the costly chain-of-thought cold-start annotations used in prior localization-based approaches. To make this cold-start-free training paradigm effective, we introduce a \textbf{re-answer} mechanism that anchors the newly acquired re-watch behavior to the model’s well-aligned prior rather than overwriting it, together with a \textbf{re-ask} mechanism that alleviates a causal-attention limitation and enables the refined answer to better attend to re-watched evidence.
By first localizing query-relevant segments and then revisiting them at higher fidelity, video-SALMONN-R$^3$ allocates its limited visual-token budget more effectively to informative regions. Experiments across six diverse benchmarks demonstrate consistent improvements over strong audio-visual baselines and prior localization-based video-LLMs. Further ablations show that these gains arise from targeted re-watching enabled by the re-answer and re-ask designs, rather than from increased token budgets or simply performing an additional answering pass. 
%We hope this work provides a simple, scalable, and reproducible framework for efficient fine-grained video understanding.

% \medskip

{\small
\bibliographystyle{unsrt}
\bibliography{neurips_2026}
}

%%%%%%%%%%%%%%%%%%%%%%%%%%%%%%%%%%%%%%%%%%%%%%%%%%%%%%%%%%%%

\appendix

\section{More Implementation Details}\label{app:prompt}

\subsection{System Prompt}

The reinforcement learning stage of video-SALMONN-R$^3$ relies on a carefully crafted system prompt that specifies the protocol in a machine-parseable schema. Rather than distilling chain-of-thought trajectories from a teacher model and cold-starting the policy via SFT, we provide only a minimal format prior through the system message below, and let the model explore full trajectories under rule-based rewards. The exact system message used in both training and inference is shown in Fig.~\ref{fig:system_prompt}.

\begin{figure}[ht]
\centering
\begin{tcolorbox}[colback=gray!5, colframe=gray!60, boxrule=0.5pt, arc=2pt, left=6pt, right=6pt, top=4pt, bottom=4pt]
\footnotesize\ttfamily
You are a helpful assistant.\\
\#\#\#\\
In the first round, you will be given a full video along with the question.\\
FIRST: Output your initial answer.\\
THEN: Think carefully to find a timestamp that is related to the question. Output the reasoning progress as an internal monologue enclosed within <thinking>...</thinking>. Your reasoning progress should follow the following logic:\\
The video generally describes ...(overall description of the video). The question asks about ...(specific detail requested). To answer this question, I should look for segments showing ...(what needs to be found). In the video, the time segment ... is related to this question because ...(reason). Therefore, I should locate this part as it contains the answer...\\
AT LAST: Output the located timestamp in second with the format <tool\_call>locate [start, end]</tool\_call>\\
Output format:\\
...<thinking>...</thinking><tool\_call>locate [start, end]</tool\_call>\\
\\
\#\#\#\\
In the second round, you will be additionally given a segmented video clip with higher framerate and resolution. You should directly output the final answer based on both the full video and the located clip. If you believe the previous answer was correct, repeat the initial answer; otherwise, correct it.
\end{tcolorbox}
\caption{The exact system prompt used for end-to-end RL in video-SALMONN-R$^3$. The first-round specification defines a three-part output (\emph{initial answer} $\to$ \texttt{<thinking>} reasoning $\to$ \texttt{<tool\_call>locate} call), while the second-round specification constrains the re-answer to be a clean final answer that either repeats or corrects the initial one.}
\label{fig:system_prompt}
\end{figure}

The prompt is deliberately structured to expose exactly three loci of model output. In the first pass, the \emph{initial answer} must be emitted \emph{before} any \texttt{<thinking>} or \texttt{<tool\_call>} tokens, which is what enables the re-answer mechanism to anchor $\mathbf{A}^{(1)}$ onto the baseline's well-aligned prior: since nothing precedes $\mathbf{A}^{(1)}$ except $(\mathbf{V}^{(1)}, \mathbf{Q})$, the first-pass answer is drawn from exactly the same distribution as the instruction-tuned base model and its competence is preserved by construction. The subsequent \texttt{<thinking>} block provides a free-form monologue slot for the model to articulate why a particular interval is relevant, and the terminal \texttt{<tool\_call>locate [start, end]</tool\_call>} yields a strictly parseable interval that drives the second-pass sampling. In the second pass, the prompt explicitly forbids meta-tokens and asks for a clean re-answer that either repeats $\mathbf{A}^{(1)}$ or corrects it, which pairs naturally with the $r_{\text{fmt2}}$ and $r_{\text{rev}}$ reward terms defined in the main text.

\subsection{LoRA Dropping}

A second implementation detail that proved important for stable RL on top of the instruction-tuned base model is what we refer to as \emph{LoRA dropping} following SALMONN \cite{tang2024salmonn}. During the audio-visual caption SFT stage, we attach a LoRA adapter with a relatively large scaling factor $\alpha_{\text{SFT}}=256$ so that the adapter has enough effective capacity to absorb the dense audio-visual captioning supervision and yield a strong instruction-tuned base model. When the same adapter is subsequently loaded at the end-to-end RL stage, however, we \emph{drop} its scaling factor to $\alpha_{\text{RL}}=32$ while keeping the learned low-rank matrices $A$ and $B$ unchanged. Concretely, since the effective LoRA contribution at inference time is
\begin{equation}
    \Delta W = \frac{\alpha}{r}\, B A,
\end{equation}
reducing $\alpha$ from $256$ to $32$ uniformly attenuates the SFT-induced weight update by a factor of $8$ before RL begins, without altering the direction of the learned adaptation.

This asymmetric $\alpha_{\text{SFT}}\!\gg\!\alpha_{\text{RL}}$ schedule serves two purposes. First, it mitigates the risk of overfitting carried over from the SFT stage: the large $\alpha_{\text{SFT}}$ is desirable for fitting captioning supervision but tends to sharpen the output distribution around SFT-preferred phrasings, which is harmful for the exploratory rollouts that DAPO relies on. By dropping $\alpha$ at the RL entry point, we soften the baseline policy and restore a substantial fraction of the backbone's pretrained behavior, which in turn enlarges the entropy of initial rollouts and gives the rule-based rewards a richer signal to shape. Second, the dropped adapter still retains the directional inductive bias learned during SFT, so the RL stage does not start from scratch; it starts from a softened version of the instruction-tuned base model that is simultaneously competent and sufficiently diverse. In practice, this schedule noticeably stabilizes the early phase of RL and reduces the frequency of mode-collapsed rollouts, which is consistent with the competence-preservation principle that underlies our re-answer design.

\subsection{Engineering Implementation Details}

Beyond the algorithmic design, the end-to-end RL stage of video-SALMONN-R$^3$ also benefits from a set of engineering optimizations. We summarize them below.

\textbf{Inference acceleration with vLLM.} Since each DAPO step requires a group of $G$ on-policy rollouts per question, the generation phase quickly becomes the dominant bottleneck if executed with vanilla HuggingFace decoding. We therefore integrate vLLM \cite{kwon2023efficient} as the rollout engine, which brings paged-attention-based KV-cache management and continuous batching. This is particularly beneficial for our setting, where rollouts exhibit highly variable lengths due to the two-pass re-watch protocol: short trajectories that terminate early can be evicted and replaced on the fly, while long trajectories keep streaming without padding waste.

\textbf{Training acceleration with Liger-Kernel.} For the forward-backward pass, we adopt Liger-Kernel \cite{hsu2025ligerkernel} to replace several memory- and compute-heavy operators in the Qwen3-VL backbone with fused Triton kernels, most notably the fused cross-entropy, RMSNorm, RoPE, and SwiGLU implementations. These fused kernels reduce both activation memory and kernel-launch overhead, which in our long-trajectory regime directly translates into a larger per-GPU effective batch and a shorter wall-clock time per optimization step. Combined with DeepSpeed ZeRO-1 \cite{rasley2020deepspeed} for optimizer-state sharding, the memory headroom freed by Liger-Kernel is what makes it feasible to train the full trajectory (with up to $\sim$20k visual tokens) end-to-end without resorting to pipeline or tensor parallelism.

\textbf{Colocate mode with bidirectional offloading.} Because video-SALMONN-R$^3$ adopts a strictly on-policy DAPO variant, the generation policy and the training policy must remain byte-identical at every step, which rules out the common asynchronous actor--learner setup. We therefore run the rollout engine and the trainer in colocate mode on the same set of GPUs, and carefully orchestrate their memory footprints via bidirectional CPU offloading: during the forward-backward pass, the vLLM engine (including its weights and KV cache) is offloaded to CPU so that the full GPU memory is available for training; during the generation phase, the trainer's model parameters, gradients, and optimizer states are offloaded to CPU, freeing the GPU for vLLM's paged KV cache. This ping-pong schedule, paired with ZeRO-1, allows a single A800 to accommodate both roles without the parameter duplication that a separate-GPU actor--learner split would incur, and keeps the on-policy invariant strictly satisfied.

\textbf{Per-GPU rollout accumulation to relieve video-decoding pressure.} A subtle but important bottleneck in audio-visual RL is CPU-side video decoding rather than GPU compute: every rollout requires decoding and resampling raw video frames, and naively replicating the same sample across all GPUs (the default behavior in many RL frameworks) multiplies this cost by the world size while yielding only a single effective group. To sidestep this, we reshape the data-parallel layout: instead of broadcasting one sample to all GPUs and running $G$ rollouts in parallel, each GPU independently processes one question and runs \texttt{accum\_step} sequential rollouts on it, after which the rollouts are all-gathered across ranks. The resulting effective batch size becomes the world size (i.e., the number of GPUs), and the effective group size becomes \texttt{accum\_step}, which precisely matches DAPO's group-normalized advantage formulation. This layout not only avoids redundant decoding of identical videos across ranks but also spreads the CPU decoding load uniformly over the cluster, which eliminates the CPU-bound stalls that would otherwise dominate the rollout phase.

\subsection{Resources}\label{app:res}

We use 32 A800s for audio alignment and audio-visual captioning SFT, and the model is trained for about 10 hours. For the RL stage, we used 96 A800s and trained the model for about 72 hours. However, the operation of decoding video in the second pass brings severe GPU utilization issues, which might be solved in the future by using async frameworks like verl. Without video decoding time, the model can be trained in about 40 hours.

\section{Reward Curves}\label{app:curve}

\begin{figure}[ht]
    \centering
    \includegraphics[width=\linewidth]{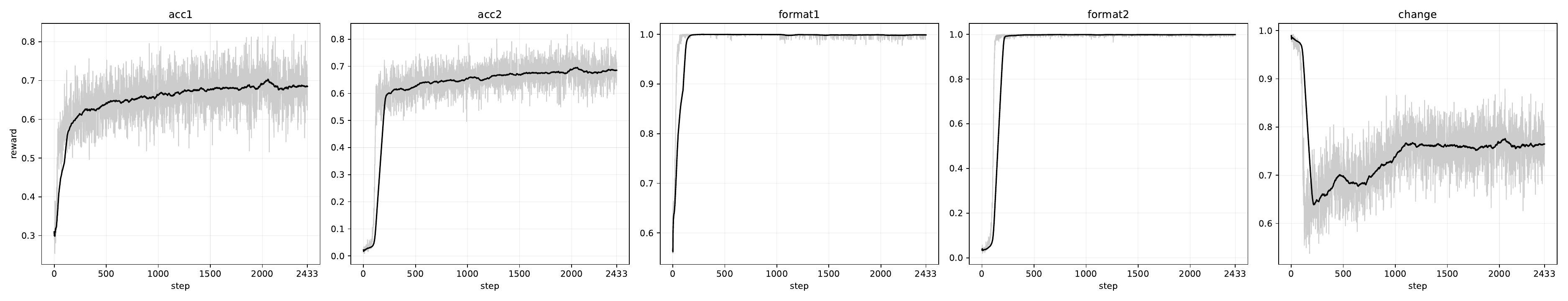}
    \caption{The reward curves in the training procedure.}
    \label{fig:curve}
\end{figure}

The reward curves are shown in Fig.~\ref{fig:curve}. In early steps, the model learns to follow the format in the instruction, while in later steps, the model learns to refine $\mathbf{A}^{(2)}$.

\section{More Ablations on RL Settings}\label{app:abl}

We further conduct additional experiments to examine the sensitivity of our recipe to key RL hyperparameters and loss formulations, with results summarized in Table~\ref{tab:rl_ablation}.

\begin{table}[ht]
    \centering
    \footnotesize
    \setlength{\tabcolsep}{4pt}
    \renewcommand{\arraystretch}{1.15}
    \caption{Ablation on RL settings. We compare different group sizes and loss formulations (GRPO vs.\ DAPO). Our default configuration (group size=8 with DAPO token-level loss) achieves the best overall performance.}
    \label{tab:rl_ablation}
    \begin{tabular}{l|cccc}
        \toprule
        \textbf{Setting} & VideoHolmes & DailyOmni & VideoMME & LVOmniBench \\
        \midrule
        group size=4        & \textbf{55.0} & 77.8 & 75.4 & 40.7 \\
        GRPO loss                      & 53.9 & 77.4 & 73.1 & 40.2 \\
        \rowcolor{gray!10}
        \textbf{Ours (group size=8, DAPO)} & 54.6 & \textbf{78.7} & \textbf{76.3} & \textbf{42.9} \\
        \bottomrule
    \end{tabular}
\end{table}

\section{Cases of How Re-Watch Helps Video Understanding}\label{app:case}

To intuitively illustrate how the re-watch mechanism contributes to fine-grained video understanding, we present several qualitative cases in Table~\ref{tab:case1}-Table~\ref{tab:case3}, each showing how the model revises the wrong initial answer and corrects to the right answer after zooming into the localized temporal segment.

\begin{table}[ht]
    \centering
    \footnotesize
    \setlength{\tabcolsep}{4pt}
    \renewcommand{\arraystretch}{1.3}
    \caption{Case 1 demonstrates the ability of removing unrelated part.}
    \label{tab:case1}
    \begin{tabular}{p{2cm}|p{2.4cm}|p{0.4cm}|p{6cm}|p{0.7cm}|p{0.4cm}}
        \toprule
        \textbf{Video Source} & \textbf{Question} & $\mathbf{A}^{(1)}$ & $\mathbf{R}^{(1)}$ & $\mathbf{T}$ & $\mathbf{A}^{(2)}$ \\
        \midrule
         VideoMME \newline FsLaTZmP6Uw & Which sentence describes this rally according to this video? A. Two athletes make a series of consecutive successful hits. B. The athlete in red can not easily make a good return. C. The athlete in black often makes edge balls. D. None of the above is correct. & B & The video generally describes a table tennis rally between two athletes, one in a red shirt and one in a black shirt, with the score displayed as CHN 3-2 JPN. The question asks about a specific detail in this rally. To answer this, I should look for segments showing the difficulty of returns by the athlete in red. In the video, the athlete in red is seen making multiple attempts to return the ball, often stretching or diving, indicating difficulty. The time segment 14.2-33.8 is related to this question because it shows the entire rally where the athlete in red struggles to return the ball effectively, while the athlete in black consistently makes powerful returns. Therefore, I should locate this part as it may be helpful. & [14.2, 33.8] & A \\
        \bottomrule
    \end{tabular}
\end{table}

\begin{table}[ht]
    \centering
    \footnotesize
    \setlength{\tabcolsep}{4pt}
    \renewcommand{\arraystretch}{1.3}
    \caption{Case 2 demonstrates the ability of locating from low fidelity to high fidelity.}
    \label{tab:case2}
    \begin{tabular}{p{2cm}|p{2.4cm}|p{0.4cm}|p{6cm}|p{0.7cm}|p{0.4cm}}
        \toprule
        \textbf{Video Source} & \textbf{Question} & $\mathbf{A}^{(1)}$ & $\mathbf{R}^{(1)}$ & $\mathbf{T}$ & $\mathbf{A}^{(2)}$ \\
        \midrule
        LVOmniBench \newline video\_72 & During the first segment of background music, by looking in the mirror, which hand can we infer the vlogger is using to hold the camera? A. Right hand. B. Left hand. C. It cannot be determined. D. Both hands. & B & The video generally describes a vlogger exploring a new home and showcasing its features. The question asks about a specific detail regarding the vlogger's hand position while looking in the mirror. To answer this question, I should look for segments showing the vlogger looking in the mirror and holding the camera. In the video, at 65s, the vlogger is seen looking in the mirror while holding the camera, and it is evident that she is using her left hand to hold the camera. & [61.3, 68.8] & A \\
        \bottomrule
    \end{tabular}
\end{table}

\begin{table}[ht]
    \centering
    \footnotesize
    \setlength{\tabcolsep}{4pt}
    \renewcommand{\arraystretch}{1.3}
    \caption{Case 3 demonstrates the ability of locating the most important clip to answer the question.}
    \label{tab:case3}
    \begin{tabular}{p{2cm}|p{2.4cm}|p{0.4cm}|p{6cm}|p{0.7cm}|p{0.4cm}}
        \toprule
        \textbf{Video Source} & \textbf{Question} & $\mathbf{A}^{(1)}$ & $\mathbf{R}^{(1)}$ & $\mathbf{T}$ & $\mathbf{A}^{(2)}$ \\
        \midrule
        VideoHolmes \newline d4Gx0UTdFiU & What is the girl's intention of adding water to nail polish in the kitchen? A. Dilute the overly thick texture. B. Test the mixing effect of liquids. C. Extend the usage time of nail polish. D. Imitate the steps of beauty tutorial. E. Pre-meditated to cover up the color of bloodstains. F. A spur-of-the-moment prank & A & The video generally describes a girl walking into a room, removing her shoes, and then sitting on the floor to apply nail polish to her feet. The question asks about her intention of adding water to the nail polish in the kitchen. To answer this question, I should look for segments showing her mixing water with nail polish. In the video, the time segment 1.5s-97.7s is related to this question because it shows the girl walking into the room, removing her shoes, and then sitting on the floor to apply nail polish to her feet. Therefore, I should locate this part as it may be helpful. & [1.5, 97.7] & E \\
        \bottomrule
    \end{tabular}
\end{table}

%%%%%%%%%%%%%%%%%%%%%%%%%%%%%%%%%%%%%%%%%%%%%%%%%%%%%%%%%%%%

\clearpage
\newpage

\end{document}